\documentclass{article} % For LaTeX2e
\usepackage{arXiv,times}

\usepackage{amsmath,amsfonts,bm}

\def\eqref#1{equation~\ref{#1}}
\def\1{\bm{1}}

\DeclareMathAlphabet{\mathsfit}{\encodingdefault}{\sfdefault}{m}{sl}
\SetMathAlphabet{\mathsfit}{bold}{\encodingdefault}{\sfdefault}{bx}{n}

\usepackage{hyperref}
\usepackage{url}
\usepackage{graphicx}
\usepackage{multirow}
\usepackage{booktabs}

\title{On multi-token prediction for efficient LLM inference}

\author{Somesh Mehra${^{1,2}}$, Javier Alonso Garcia$^1$, Lukas Mauch$^1$ \\
$^1$European Research Center (EUREC), Sony Europe B.V. \\
$^2$Swiss Federal Institute of Technology Lausanne (EPFL) \\
\texttt{somesh.mehra@epfl.ch}, \texttt{\{javier.alonsogarcia, lukas.mauch\}@sony.com}
}

\iclrfinalcopy % Uncomment for camera-ready version, but NOT for submission.
\begin{document}

\maketitle

\begin{abstract}

We systematically investigate multi-token prediction (MTP) capabilities within LLMs pre-trained for next-token prediction (NTP). We first show that such models inherently possess MTP capabilities via numerical marginalization over intermediate token probabilities, though performance is data-dependent and improves with model scale. Furthermore, we explore the challenges of integrating MTP heads into frozen LLMs and find that their hidden layers are strongly specialized for NTP, making adaptation non-trivial. Finally, we show that while joint training of MTP heads with the backbone improves performance, it cannot fully overcome this barrier, prompting further research in this direction. Our findings provide a deeper understanding of MTP applied to pretrained LLMs, informing strategies for accelerating inference through parallel token prediction.

\end{abstract}

\section{Introduction}
In recent years, decoder-only transformers  have emerged as the state-of-the-art models for language modeling and are widely adopted for large language models (LLMs). However, as these models grow in size and complexity, so do the challenges associated with their inference, due to the inherent sequential and memory bandwidth bounded nature of this process~\citep{memory-bandwidth-bound-squeezellm}.

In this context, there is growing interest in multi-token prediction (MTP) methods that diminish the purely auto-regressive nature of LLMs, enabling them to generate multiple adjacent tokens in parallel for a given partial input sequence. This can be considered a form of temporal sparsity, whereby a single forward pass is reused to predict multiple tokens, thus reducing sequential dependencies.
MTP has shown great promise in accelerating inference through self-speculative decoding, which in practice can yield speedups of up to $3.6\times$ \citep{medusa}. 

While previous works have shown the feasibility of MTP, there is still a lack of understanding about the pre-requisites and limitations of MTP when applied to modern LLMs that are pre-trained for next-token prediction (NTP). Works such as~\citet{meta-mtp} and~\citet{mtp-tensor-decomp} train MTP models from scratch, primarily using it as a tool to improve NTP performance. \citet{medusa} introduce MEDUSA, a framework for attaching MTP heads to a pretrained LLM, however they use it only for accelerating inference, and the underlying MTP performance is hidden (only speedups are reported).

This paper addresses this knowledge gap by investigating MTP adaptation of NTP models. Specificaly, we make the following contributions:

{\bf MTP capabilities of NTP models}: We demonstrate that LLMs trained only for NTP already possess MTP capabilities, i.e. MTP is possible through numeric marginalization over intermediate token probabilities. Although this is computationally complex, it is a strong performance baseline.

{\bf Model size dependence}: We observe that the MTP capabilities of LLMs are strongly data dependent and improve with increasing model size.

{\bf Challenges in adding MTP heads}: Attaching additional MTP heads to a frozen LLM backbone is not straightforward, as hidden layers tend to specialize for NTP tasks.

{\bf Joint training strategies}: We explore joint training of the LLM backbone and MTP heads, which can improve performance but still falls short of the baseline, warranting further study.

\section{Multi-token prediction capabilities of transformers trained for next-token prediction}
We delve into the multi-token prediction (MTP) capabilities of transformers trained for next-token prediction (NTP), leveraging simple information-theoretic criteria to gain deeper insights. Specifically, let 
$\mathcal{X} = \{x_1, x_2, ..., x_T \} \sim p(\mathcal{X})$ be a token sequence from the true data generating distribution $ p(\mathcal{X})$, where $x_t \in \mathbb{V}$ with vocabulary $\mathbb{V}$. Further, let $\mathcal{X}_{\leq t} = \{x_1, x_2, ..., x_t \}$ and $\mathcal{X}_{t:t+k} = \{x_{t+1}, x_{t+2}, ..., x_{t+k} \}$ denote sub-sequences. A transformer trained for NTP learns the factorization 
\begin{align}
    p(\mathcal{X}; \theta) = p(x_1) \prod_{t=1}^{T-1} p(x_{t+1}|\mathcal{X}_{\leq t}; \theta)\;,
    \label{eq:auto_regressive_fact}
\end{align}
where $\theta$ are the model parameters. NTP samples from $p(\mathcal{X}; \theta)$ in an auto-regressive way, i.e. generating all $x_{t+1} \sim p(x_{t+1}|\mathcal{X}_{\leq t}; \theta), \:\: \forall t=1,...,T-1$ sequentially.

On the other hand, MTP means to sample a full sub-sequence of length $K$ tokens given the context $\mathcal{X}_{\leq t}$ in parallel
\begin{align}
    \mathcal{X}_{t:t+K} \sim p(\mathcal{X}_{t:t+K}|\mathcal{X}_{\leq t}; \theta)\;.
\end{align}
Of course, this is intractable for the given auto-regressive factorization given in Eq.~\ref{eq:auto_regressive_fact}. However, assuming conditional independence of the tokens within $\mathcal{X}_{t:t+K}$ given $\mathcal{X}_{\leq t}$, we can compute $p(\mathcal{X}_{t:t+K}|\mathcal{X}_{\leq t}; \theta)$, using marginalization:
\begin{align}
    p(\mathcal{X}_{t:t+K}|\mathcal{X}_{\leq t}; \theta) = p(x_{t+1}|\mathcal{X}_{\leq t}; \theta) \prod_{k=2}^K \sum_{s_{1:k-1} \in \mathbb{V}^{k-1}} p(x_{t+k}, s_{1:k-1}|\mathcal{X}_{\leq t}; \theta)\;.
    \label{eq:marginalization}
\end{align} 
% \begin{align}
%     p(x_{t+k}, s_{1:k-1}|\mathcal{X}_{\leq t}; \theta)
%     &= p(x_{t+k} | ( \mathcal{X}_{\leq t}, s_{1:k}) ; \theta) \prod^{k-1}_{i=1} p(s_{k-i} | ( \mathcal{X}_{\leq t}, s_{1:k-i-1}) ; \theta)
% \end{align}
Note that in practice, this is only feasible for small $k$ because of the sum over all possible subsequences of length $K-1$.

We evaluate MTP using the marginalization in Eq.~\ref{eq:marginalization} across different model families\footnote{EleutherAI's Pythia model suite ~\citep{pythia} and Meta's Llama-3 models~\citep{llama3}.} and sizes using $K=2$. The details of all our experiments are summarized in Appendix~\ref{ap:exp_details}. Figure~\ref{fig:mtp_performance_marginalization} shows the obtained top-5 token accuracy (higher is better) for open-ended generation and translation. We see, that: 1)~LLMs trained for NTP can perform MTP. 2)~MTP capabilities grow with model size. 3)~Performance is strongly data dependent.

The improvement with scale can be attributed to a sparsification of next-token probabilities in larger models, indicating that they become more selective in their possible completions, simplifying marginalization (see Appendix~\ref{ap:sparsification-of-ntp-probs} for details). The data dependence of MTP performance is also not surprising: for tasks such as translation, the provided context (i.e. the sentence in the source language) reduces ambiguity in predicting $\mathcal{X}_{t:t+K}$, making the predictions more deterministic.

\begin{figure}[!t]
\begin{center}
\includegraphics[width=0.7\linewidth]{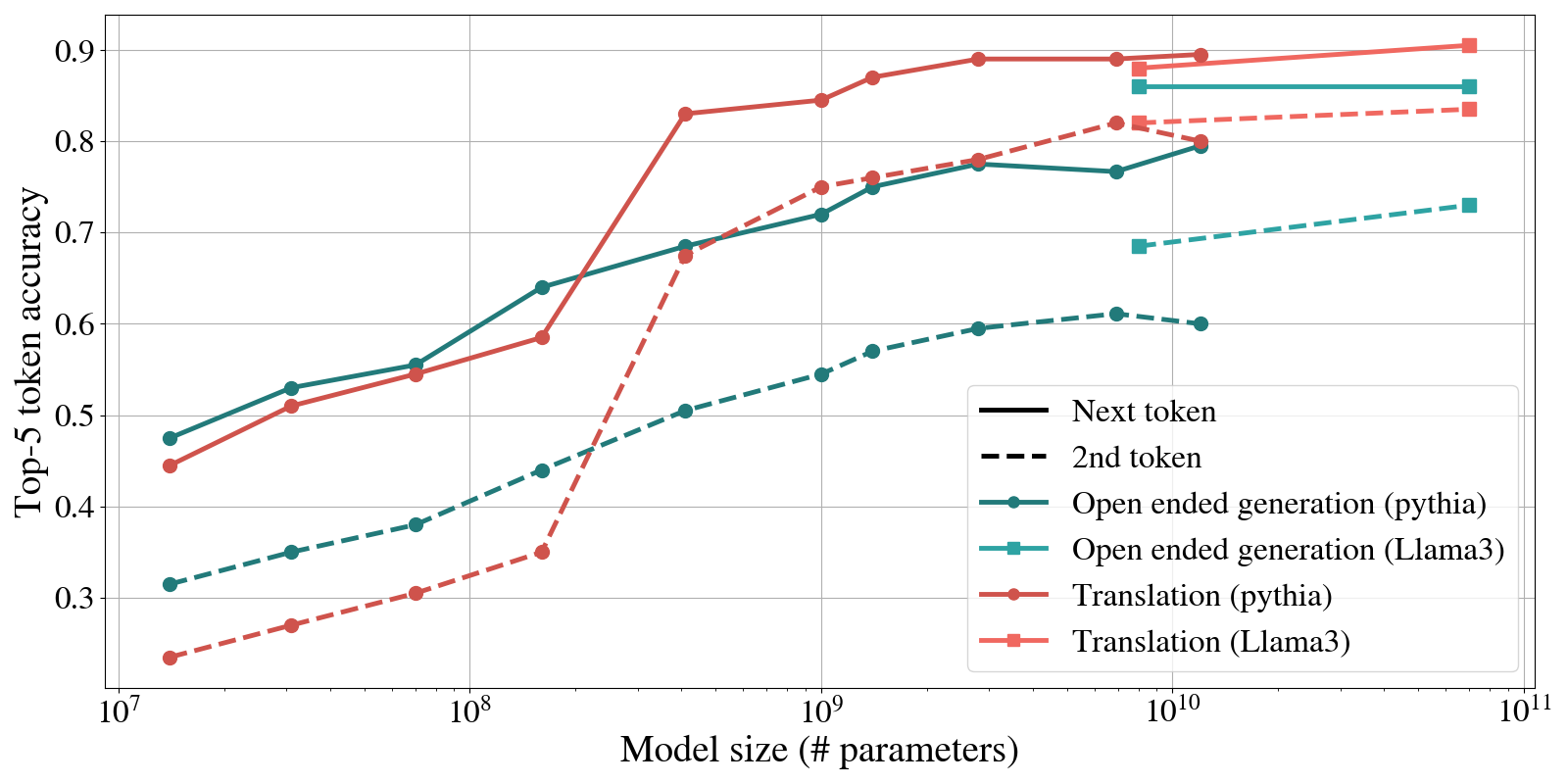}
\end{center}
\vspace{-10px}
\caption{Top-5 accuracy of MTP using marginalization, for open-ended generation and translation across model families and sizes. MTP capabilities grow with model size and are data dependent.}
\label{fig:mtp_performance_marginalization}
\end{figure}

\section{Finetuning transformers for multi-token prediction}

Previous works such as MEDUSA adapt LLMs from NTP to MTP by attaching MTP heads to a pretrained model to amortize the marginalization from Eq.~\ref{eq:marginalization}. We take a similar approach with some key modifications: rather than attaching additional heads on top of the pretrained LLM, we replicate its final transformer layer N times to create N independent heads within the model. Each head processes the final hidden state from the shared backbone and applies a shared unembedding layer to compute token probabilities, avoiding the inefficiency of training separate per-token unembeddings.

More formally, let $L$ be the number of transformer layers in the original LLM and $h$ be the hidden size of the model. Let $z^{L-1}_{1:t} \in \mathbb{R}^{h\times t}$ be the hidden states from the penultimate transformer layer of the original model given the input sequence $\mathcal{X}_{\leq t}$. Define $f^L_n:\mathbb{R}^{h} \to \mathbb{R}^{h}$ as the final transformer layer for the $n^\text{th}$ head, and $f_u:\mathbb{R}^h \to \mathbb{R}^{|\mathbb{V}|}$ as the unembedding to map hidden representations to logits. The $n^\text{th}$ token probabilities are thus given by 
\begin{align}
    p(x_{t+n}|\mathcal{X}_{\leq t}; \theta) = \sigma \big( f_u( f_n^L(z^{L-1}_{1:t})) \big) \:,
\end{align}
where $n\in \{1,2,...N\}$ and $\sigma(\cdot)$ is the softmax function. Further details and a visualization of this architecture are available in Appendix~\ref{ap:mtp_finetuning}. 

To simplify analysis, we use $N=2$. We train the heads, with the backbone and unembedding frozen, on a translation task due to its suitability for MTP. Full training details are outlined in Appendix~\ref{ap:exp_details}.

Our experiments show that this approach is not straightforward due to an early specialization of the LLM for NTP. To visualize this specialization, we compute the Kullback-Leibler (KL) divergence between intermediate token probabilities and the final token probabilities. 
More specifically, let $z^{\ell}_t \in \mathbb{R}^{h} \text{ with } \ell=1,2,...,L-1$ be the output of layer $\ell$ at sequence position $t$. We compute the intermediate token probabilities $p_\ell(x_{t+1}|\mathcal{X}_{\leq t}; \theta)$ by applying the final unembedding layer directly to $z^{\ell}_t$, i.e. $p_\ell(x_{t+1}|\mathcal{X}_{\leq t}; \theta) = \sigma(f_u ( z^{\ell}_t))$ (depicted visually in Appendix~\ref{ap:mtp_finetuning}). Using this, we can calculate $\mathrm{KL}\big(p_\ell(x_{t+1}|\mathcal{X}_{\leq t}; \theta) \,||\, p(x_{t+1}|\mathcal{X}_{\leq t}; \theta)\big)$.

\begin{figure}[!b]
\begin{center}
\includegraphics[width=0.7\linewidth]{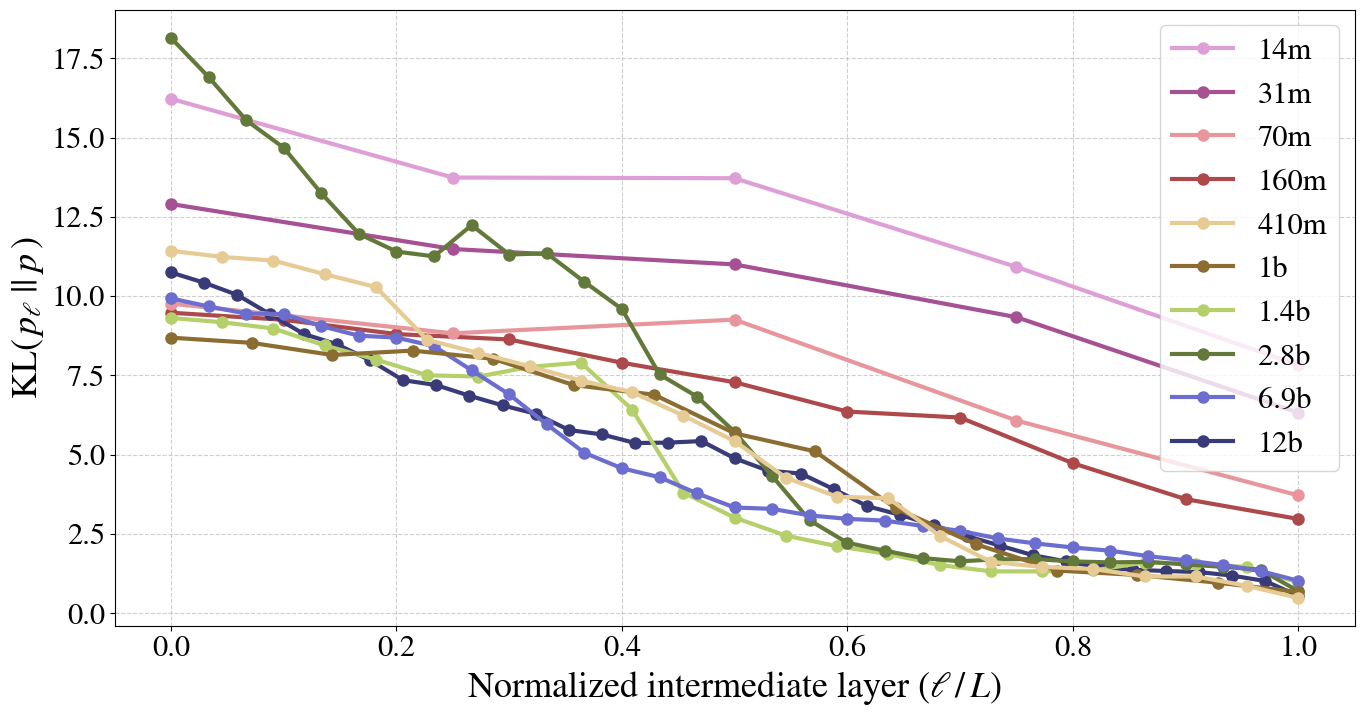}
\end{center}
\vspace{-9px}
\caption{KL divergence between intermediate and final token probabilities. For large models, intermediate layers reach a representation close to the final output relatively early, indicating strong specialization to NTP.}
\label{fig:early_specialization}
\end{figure}

As shown in Figure~\ref{fig:early_specialization}, we observe that particularly in large models, intermediate layers start specializing for NTP even before reaching the output layer. This means that the network has likely discarded information crucial for MTP, making it difficult to achieve good results by simply adding and finetuning MTP heads. 

Thus, there is strong motivation for jointly finetuning the backbone and MTP heads to try  to overcome this specialization and improve MTP performance. \citet{medusa} also report speedups with joint finetuning in MEDUSA, and propose three strategies for joint training: 1)~\textit{Balancing NTP and MTP losses}, to train the heads while preserving NTP capabilities; 2)~\textit{Differential learning rates}, where heads are trained with a higher learning rate for faster convergence; and 3)~\textit{Head warmup}, training the heads with a frozen backbone before joint training.

For the first strategy, we rescale each head's loss to normalize the Root Mean Squared (RMS) losses to that of the first head (per batch). We conduct an ablation study for the other two strategies to determine which is most effective. The backbone is finetuned using LORA~\citep{lora} with adapters of rank 8 applied to all query, key, and value layers, while all independent head parameters are trained in all MTP experiments. For a baseline comparison, we also finetune the original model with the same LORA configuration on the same task, and use the marginalization from Eq.~\ref{eq:marginalization} to estimate MTP capabilities. Our analysis focuses on a subset of Pythia models that exhibit strong MTP capabilities and are feasible within our training budget (410M–2.8B parameters).

As shown in Table~\ref{tab:results}, joint finetuning consistently improves MTP performance over training only the heads. The proposed strategies also outperform vanilla joint finetuning, though no single combination is clearly most effective. Yet, a significant performance gap remains compared to the baseline.

To address the challenge of breaking NTP specialization in pretrained LLMs, we propose using weighted hidden states (WHS). Instead of using the final output of the backbone as input to the heads, we take a weighted sum of all intermediate hidden states from the backbone, aiming to leverage earlier layers' information before specialization. Formally, the $n^\text{th}$ token probabilities become: 
\begin{align}
    p(x_{t+n}|\mathcal{X}_{\leq t}; \theta) = \sigma \bigg( f_u \big(f_n^L(\sigma(W^n_{s} / T) \cdot Z^{1:L-1}_{1:t})\big) \bigg) \:,
\end{align}
where $Z^{1:L-1}_{1:t}\in \mathbb{R}^{(L-1) \times h \times t}$ are the hidden states of each layer in the backbone, $W^n_{s} \in \mathbb{R}^{(L-1)}$ are the learnable weights of each intermediate layer for head $n$, and $T$ is a fixed temperature for scaling the weights before normalizing with softmax. This is again depicted visually in Appendix~\ref{ap:mtp_finetuning}.

\begin{table}[!t]
\caption{Top-5 next (left) and 2nd (right) token prediction accuracies for different joint finetuning strategies. Compared to training the heads only (i.e. with frozen backbone), 2nd token prediction consistently improves with joint finetuning, and using weighted hidden states with a differential LR appears to be a promising strategy to improve performance.}
    \label{tab:results}
\begin{center}
    \begin{tabular}{l||*{3}{c@{\hspace{6pt}}c|}c@{\hspace{6pt}}c}
        \hline
        & \multicolumn{2}{c|}{410m} & \multicolumn{2}{c|}{1b} & \multicolumn{2}{c|}{1.4b} & \multicolumn{2}{c}{2.8b} \\
        \hline
        \color{gray}Baseline (FT base model) & \color{gray}0.827 & \color{gray}0.682 & \color{gray}0.847 & \color{gray}0.730 & \color{gray}0.866 & \color{gray}0.760 & \color{gray}0.881 & \color{gray}0.777 \\
        Heads only & 0.798 & 0.500 & 0.816 & 0.590 & 0.845 & 0.573 & 0.856 & 0.607 \\
        \hline
        \hline
        - & 0.742 & 0.538 & \textbf{0.812} & 0.603 & 0.825 & 0.608 & 0.828 & 0.645 \\
        Head warmup & \textbf{0.766} & 0.566 & 0.810 & 0.621 & 0.838 & 0.620 & \textbf{0.839} & 0.650 \\
        Diff. LR & 0.759 & 0.562 & 0.809 & 0.618 & 0.829 & 0.629 & 0.836 & \textbf{0.667 }\\
        Head warmup + Diff. LR & 0.755 & 0.572 & \textbf{0.812} & 0.615 & \textbf{0.844} & 0.633 & 0.835 & 0.654 \\
        \hline
        Diff. LR + WHS & 0.765 & \textbf{0.590} & 0.807 & \textbf{0.623} & 0.831 & \textbf{0.642} & 0.833 & 0.659 \\
        \hline
    \end{tabular}
    \vspace{-10px}
\end{center}
\end{table}

We perform joint finetuning using WHS combined with the differential LR strategy to accelerate convergence (full experiment details in Appendix~\ref{ap:exp_details}). Results in Table~\ref{tab:results} show WHS is a promising approach, achieving the overall best MTP performance among the tested strategies.

Still, the performance gap compared to the baseline remains. Intuitively however, this makes sense. A good NTP model that specializes strongly is likely to be more confident in its token predictions, which by our analysis means it can better perform MTP. However, this same phenomenon makes MTP challenging to amortize, since the model has not learned a good shared representation for MTP, which is a non-trivial challenge to overcome. We discuss possible future directions in Appendix~\ref{future-directions}.

\section{Conclusion}

Our research shows that while NTP models can perform well on MTP tasks, amortizing this capability with MTP heads is challenging. Strong NTP specialization enables effective MTP via marginalization but makes adaptation difficult. Even so, MTP adaptation remains useful in scenarios requiring only rough token drafts (e.g., self-speculative decoding). While training the strongest MTP models likely requires full MTP pretraining, there is still room to improve MTP adaptation of NTP models. Given its effectiveness, despite the computational cost, we recommend using marginalization-based approaches as a baseline for future research.

\bibliography{arXiv}

\begin{thebibliography}{8}
\providecommand{\natexlab}[1]{#1}
\providecommand{\url}[1]{\texttt{#1}}
\expandafter\ifx\csname urlstyle\endcsname\relax
  \providecommand{\doi}[1]{doi: #1}\else
  \providecommand{\doi}{doi: \begingroup \urlstyle{rm}\Url}\fi

\bibitem[AI@Meta(2024)]{llama3}
AI@Meta.
\newblock Llama 3 model card.
\newblock 2024.
\newblock URL \url{https://github.com/meta-llama/llama3/blob/main/MODEL_CARD.md}.

\bibitem[Basharin et~al.(2024)Basharin, Chertkov, and Oseledets]{mtp-tensor-decomp}
Artem Basharin, Andrei Chertkov, and Ivan Oseledets.
\newblock Faster language models with better multi-token prediction using tensor decomposition.
\newblock \emph{arXiv preprint arXiv:2410.17765}, 2024.

\bibitem[Biderman et~al.(2023)Biderman, Schoelkopf, Anthony, Bradley, O’Brien, Hallahan, Khan, Purohit, Prashanth, Raff, et~al.]{pythia}
Stella Biderman, Hailey Schoelkopf, Quentin~Gregory Anthony, Herbie Bradley, Kyle O’Brien, Eric Hallahan, Mohammad~Aflah Khan, Shivanshu Purohit, USVSN~Sai Prashanth, Edward Raff, et~al.
\newblock Pythia: A suite for analyzing large language models across training and scaling.
\newblock In \emph{International Conference on Machine Learning}, pp.\  2397--2430. PMLR, 2023.

\bibitem[Cai et~al.(2024)Cai, Li, Geng, Peng, Lee, Chen, and Dao]{medusa}
Tianle Cai, Yuhong Li, Zhengyang Geng, Hongwu Peng, Jason~D. Lee, Deming Chen, and Tri Dao.
\newblock Medusa: Simple llm inference acceleration framework with multiple decoding heads.
\newblock \emph{arXiv preprint arXiv: 2401.10774}, 2024.

\bibitem[Cettolo et~al.(2017)Cettolo, Federico, Bentivogli, Niehues, St{\"u}ker, Sudoh, Yoshino, and Federmann]{iwslt-2017}
Mauro Cettolo, Marcello Federico, Luisa Bentivogli, Jan Niehues, Sebastian St{\"u}ker, Katsuhito Sudoh, Koichiro Yoshino, and Christian Federmann.
\newblock Overview of the {IWSLT} 2017 evaluation campaign.
\newblock In \emph{Proceedings of the 14th International Conference on Spoken Language Translation}, pp.\  2--14, Tokyo, Japan, December 14-15 2017. International Workshop on Spoken Language Translation.
\newblock URL \url{https://aclanthology.org/2017.iwslt-1.1}.

\bibitem[Gloeckle et~al.(2024)Gloeckle, Idrissi, Rozière, Lopez-Paz, and Synnaeve]{meta-mtp}
Fabian Gloeckle, Badr~Youbi Idrissi, Baptiste Rozière, David Lopez-Paz, and Gabriel Synnaeve.
\newblock Better \& faster large language models via multi-token prediction, 2024.
\newblock URL \url{https://arxiv.org/abs/2404.19737}.

\bibitem[Hu et~al.(2021)Hu, Shen, Wallis, Allen-Zhu, Li, Wang, Wang, and Chen]{lora}
Edward~J Hu, Yelong Shen, Phillip Wallis, Zeyuan Allen-Zhu, Yuanzhi Li, Shean Wang, Lu~Wang, and Weizhu Chen.
\newblock Lora: Low-rank adaptation of large language models.
\newblock \emph{arXiv preprint arXiv:2106.09685}, 2021.

\bibitem[Kim et~al.(2023)Kim, Hooper, Gholami, Dong, Li, Shen, Mahoney, and Keutzer]{memory-bandwidth-bound-squeezellm}
Sehoon Kim, Coleman Hooper, Amir Gholami, Zhen Dong, Xiuyu Li, Sheng Shen, Michael~W Mahoney, and Kurt Keutzer.
\newblock Squeezellm: Dense-and-sparse quantization.
\newblock \emph{arXiv preprint arXiv:2306.07629}, 2023.

\end{thebibliography}
\bibliographystyle{arXiv}

\appendix
\section{Experiment details}
\label{ap:exp_details}

\subsection{Marginalization}
\label{ap:exp_details_marginalization}

As per the formulation in Eq.~\ref{eq:marginalization}, to estimate $2^{\text{nd}}$ token probabilities using a NTP model, we can marginalize over all possible $x_{t+1}$ to calculate 
\begin{align}
    p(x_{t+2}|\mathcal{X}_{\leq t}; \theta) = \sum_{y \in \mathbb{V}} p(x_{t+2}|\mathcal{X}_{\leq t}, y; \theta) \cdot p(y|\mathcal{X}_{\leq t}; \theta)\;,
    \label{eq:2nd-token-marginalization}
\end{align}
where $\mathbb{V}$ is the vocabulary set of the model. 
Given the large size of $|\mathbb{V}|$ in LLMs however, in practice we only consider the set of tokens which are contained in the top 0.99 of the predicted probability distribution $p(x_{t+1}|\mathcal{X}_{\leq t}; \theta)$, and then normalize the resulting distribution to sum to 1. 

\subsection{Datasets}
\label{ap:datasets}
Experiments are primarily conducted using the IWSLT 2017 translation dataset~\citep{iwslt-2017}, specifically using the de-en language direction. The dataset consists of pairs of sentences in the source language (German), and the same sentence translated into the target language (English). Using this, we construct a single sequence using the following template:

\begin{quote}
    \textit{Translate the following German sentence to English: [source sentence]\\ -\--\-- \\ Translated: [target sentence]}
\end{quote}

For all trainings (MTP models and finetuning of base NTP models), we use 200k samples from the train set, and calculate losses only over the target sentence. 

For evaluation, since we are comparing with marginalized results as a baseline (which is slow to compute), we evaluate over 50 validation samples for the trained MTP model experiments, and 10 validation samples for the initial marginalization experiments (reported in Figure~\ref{fig:mtp_performance_marginalization}), since this analysis scales to very large models and would be otherwise infeasible. For each validation sample, we calculate top-5 accuracies (as outlined in \ref{ap:exp-details-evaluation}) for the last 20 tokens in the sequence. The validation samples are filtered such that the translated sentence is $\geq 20$ tokens to ensure that models are only evaluated on the translation task.

For the open-ended generation results reported in Figure~\ref{fig:mtp_performance_marginalization}, we follow the same setup but instead use 10 reference generations from GPT-4o, where we prompted the model to generate sequences of approximately 200 tokens about any topics (to ensure that there is enough context for the model to predict multiple tokens independently).

\subsection{Evaluation}
\label{ap:exp-details-evaluation}

All next and $2^{\text{nd}}$ token top-5 accuracy results are calculated as the proportion of tokens from a reference sequence that are contained in the top-5 of the respective predicted probability distributions. For marginalization experiments with NTP models, these are the actual next token probabilities and the estimated $2^\text{nd}$ token probabilities (as per equation \ref{eq:2nd-token-marginalization}). For MTP models, these are the probability distributions outputted from each head. 

\subsection{Training}

All trainings are performed for 1 epoch using a cross entropy loss over the translation dataset described in Appendix~\ref{ap:datasets}. An effective batch size of 64 is used in all cases, along with a Cosine Annealing learning rate scheduler and AdamW optimizer with 0.01 weight decay. Each model is trained with a learning rate equal to $0.5\times$ the learning rate it was pretrained with. The learning rates for each of the models we train is shown in Table~\ref{tab:training-hyperparams}. 

\begin{table}[h]
    \caption{Learning rates used for Pythia models of different sizes.}
    \label{tab:training-hyperparams}
    \begin{center}
        \begin{tabular}{c|c}
            \hline
            Model size & LR  \\
            \hline
            410M & $1.5e^{-4}$  \\
            1B   & $1.5e^{-4}$  \\
            1.4B & $8e^{-5}$  \\
            2.8B & $6e^{-6}$  \\
            \hline
        \end{tabular}
    \end{center}
\end{table}

{\bf Heads only: }each head is optimized independently since all shared parameters (backbone and unembedding) are frozen. 

{\bf Vanilla joint finetuning: }LORA adapters of rank 8 are applied to all query, key and value layers in the backbone. The shared unembedding remains frozen. The $N$ independent heads are fully trainable (i.e. all parameters in the respective transformer layers are trained). The first head is initialized to the final transformer layer of the original model, while the remaining heads are randomly initialized.

{\bf Joint finetuning with head warmup: }uses the same settings as vanilla joint finetuning, however the heads are initialized to the heads obtained after 1 epoch of frozen backbone training.

{\bf Joint finetuning with differential learning rate: }uses the same settings as vanilla joint finetuning, however the independent heads use a $4\times$ higher learning rate than the base learning rate.

{\bf Joint finetuning with weighted hidden states: }uses the same settings as the joint finetuning with differential learning rate, except that the head inputs are modified as described in Appendix~\ref{ap:mtp_finetuning}. We use a temperature $T=0.1$, and uniformly initialize the layer weights to 0.1.

{\bf LORA finetuning of base model: }we apply the same settings as above when finetuning the base model on translation to use as a baseline, except that no layers are fully trainable, and the LORA adapters (with rank 8) are attached to all query, key and value layers throughout the model.

\section{Sparsification of NTP probabilities}
\label{ap:sparsification-of-ntp-probs}

\begin{figure}[h] 
\begin{center}
\includegraphics[width=0.7\linewidth]{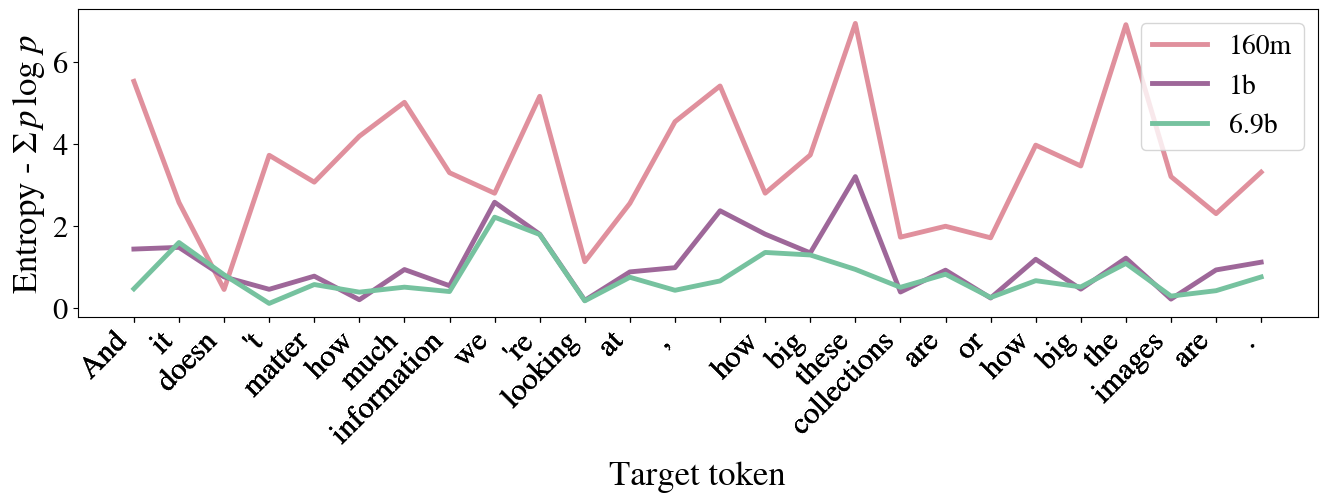}
\end{center}
\caption{Entropy of token probabilities over an example translated sequence for various Pythia models. Larger models tend to be more confident in NTP across majority of the sequence.}
\label{fig:entropy_visualization}
\end{figure}

\begin{figure}[h] 
\begin{center}
\includegraphics[width=0.7\linewidth]{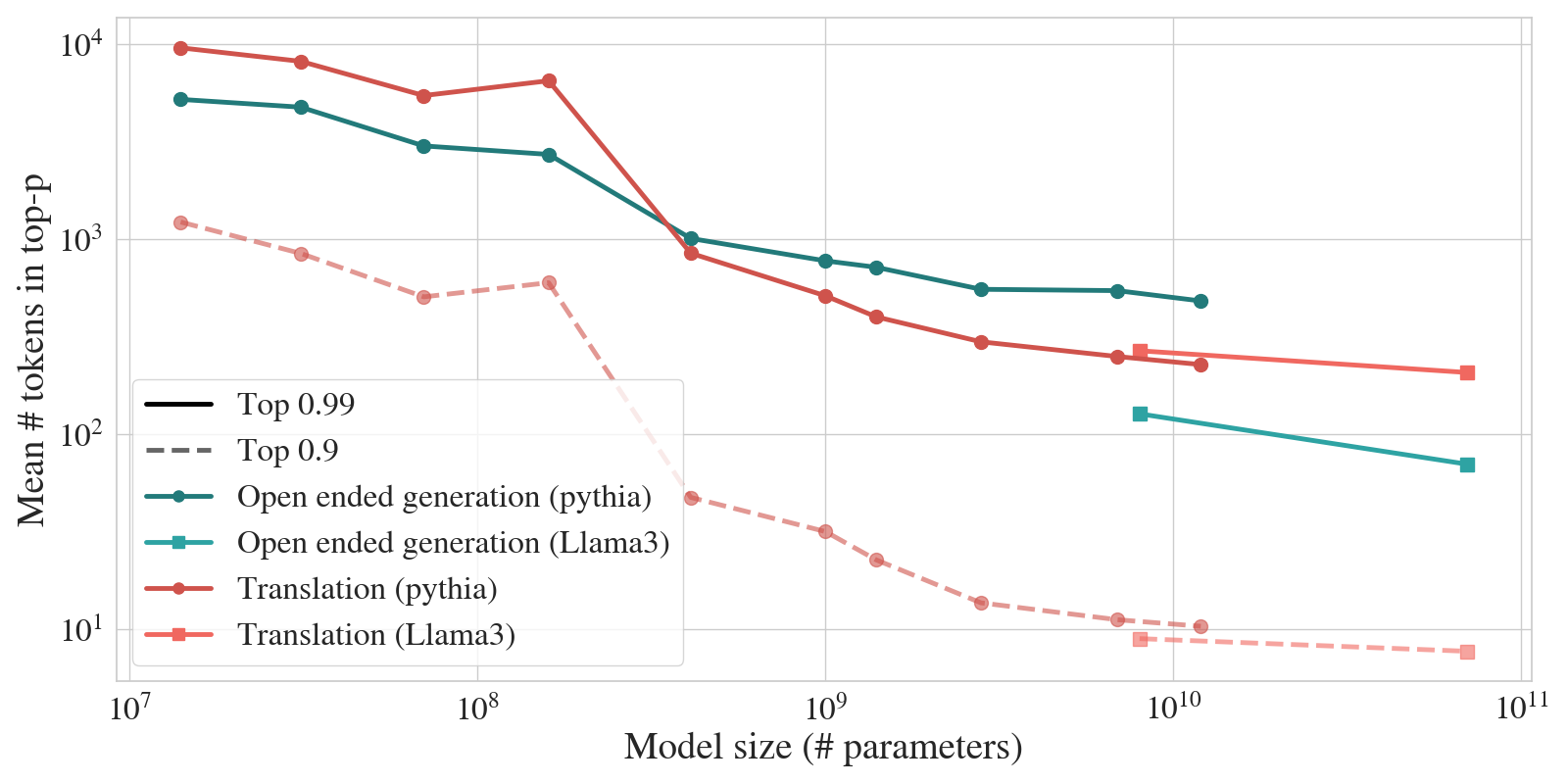}
\end{center}
\caption{Average number of considered $x_{t+1}$ tokens -- i.e. tokens in the top 0.99 (solid) of the predicted probability distribution -- for each model size and task during the marginalization analysis. This number consistently decreases with model size, and comparing to the number of tokens in the top-0.9 (dashed) of the distribution further indicates that many of these considered tokens would also have very low probability.}
\label{fig:tokens_in_top_p}
\end{figure}

To understand why LLMs are able to perform MTP better with increasing model size, we investigate the predicted probability distributions of these models. We find that this is likely caused by a sparsification of the NTP probabilities $p(x_{t+1}|\mathcal{X}_{\leq t}; \theta)$ as model size increases. More specifically, larger models tend to be very selective on the possible completions for a given context $\mathcal{X}_{\leq t}$, which makes marginalization simpler. Figure~\ref{fig:entropy_visualization} visualizes this sparsification for an example sequence, as the entropy of the token probabilities decreases with growing model size. Figure~\ref{fig:tokens_in_top_p} reinforces this idea, as the number of tokens in the top-0.99 of the probability distribution decreases with model size, indicating that the probability mass is concentrated on fewer tokens.

\section{MTP model setup}
\label{ap:mtp_finetuning}

\begin{figure}[h]
\begin{center}
\includegraphics[width=1\linewidth]{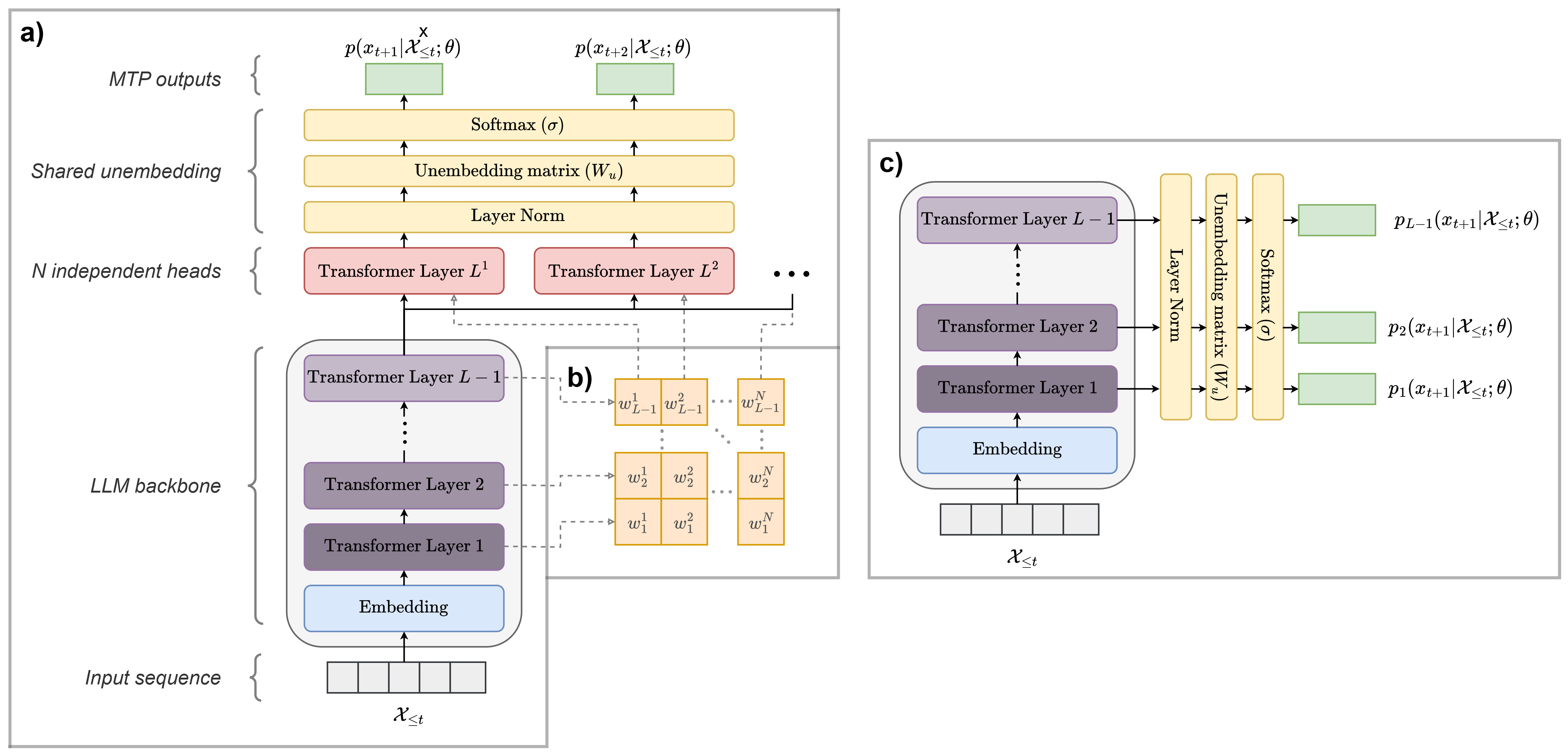}
\end{center}
\caption{\textbf{a)} Overview of the MTP model architecture used for experimentation. The model consists of a LLM backbone, $N$ independent heads which take as input the final hidden states from the backbone, and a shared unembedding applied to each head. Thus, for a given input sequence, we predict $N$ token probabilities with a single forward pass through the backbone. \textbf{b)} When using weighted hidden states, instead of taking the final output of the backbone as input to each head, we instead take the weighted sum of all intermediate hidden states from the backbone. Each head learns its own weight vector, which is normalized before taking the weighted sum. \textbf{c)} Intermediate token probabilities are calculated by taking the hidden states from each layer in the backbone and applying the shared unembedding to obtain token probabilities.}
\label{fig:mtp_model_architecture}
\end{figure}

This section gives the details about our architecture for multi-token prediction. The vanilla MTP model architecture is depicted in Figure~\ref{fig:mtp_model_architecture}a. The LLM backbone is directly taken from the pretrained base NTP model we are adapting (minus the final transformer layer). Each independent head consists of a single transformer layer with the same structure as in the backbone. The first head is initialized to the final transformer layer of the base model (i.e. the output of the first head will be initially equivalent to the original model), whilst all additional heads are randomly initialized. The shared unembedding is applied on top of each head, and consists of a layer normalization, an unembedding matrix $W_u \in \mathbb{R}^{|\mathbb{V}|\times h}$, and a softmax operation, to obtain token probabilities from each head. This unembedding is again taken from the pretrained base model and is frozen for all experiments. 

The resulting architecture has the same structure as that proposed by~\citet{meta-mtp} i.e. a shared LLM backbone, N independent transformer layers, and a shared unembedding. However instead of training from scratch, we use a pretrained NTP model to obtain the backbone and unembedding.

MEDUSA~\citep{medusa} on the other hand attaches additional linear heads to the final transformer block of the pretrained backbone, each including its own unembedding matrix. Although this work is the closest to ours in terms of adapting NTP models for MTP, we deviate from this architecture for the following reasons: 
\begin{enumerate}
    \item Using a shared unembedding saves significant resources (both memory and computation) during training and inference, since these matrices are large (order $|\mathbb{V}|$). Using the pretrained unembedding also allows this matrix to be fixed during training, drastically reducing the number of trainable parameters especially as the number of heads scales. 
    \item Using transformer layers instead of linear layers allows the heads to attend to previous tokens in the sequence, rather than simply probing the final hidden representation for a single token position.
    \item Instead of attaching additional transformer layers atop the full pretrained backbone, we attach them one layer below, in order to maintain the same depth as the original model.
\end{enumerate}

\section{Future directions}
\label{future-directions}

Based on our findings, we propose the following as a few research directions which could potentially improve MTP adaptation of NTP models:

{\bf Further exploration of weighted hidden states: }in this work, we perform some preliminary experiments using weighted hidden states, with promising initial results. However, there is a lot of scope to further explore this idea, and various hyperparameters for the weights, such as using better initialization and normalization schemes or increasing the learning rate (to facilitate faster convergence)  could improve performance. 

{\bf Multi-layer MTP adaptation: }our MTP architecture uses only 1 independent transformer layer per head, taking the output of the $L-1^{\text{th}}$ layer of the original model as input, by which point the hidden states have likely already specialized. Increasing the number of layers that we use for MTP adaptation however could be beneficial, since at the same time we (somewhat) mitigate the specialization problem by taking the outputs from earlier, less specialized layers, whilst also enabling more powerful MTP heads which can learn better adaptations for $n^{\text{th}}$-token prediction. Of course, as we increase the number of layers, the additional parameters also scale (by $N \times L_{h}$ where $L_{h}$ is the number of layers we use per head), and eventually we would simply degrade to training $N$ complete LLMs each for $n^{\text{th}}$-token prediction. Thus, a balance is required, and likely only a small number of layers is feasible before it becomes more beneficial to simply train an MTP model from scratch.

\end{document}